\documentclass[runningheads]{llncs}
\usepackage{float}
\usepackage{graphicx}
\usepackage[normalem]{ulem}
\usepackage{multirow}
\useunder{\uline}{\ul}{}
\begin{document}
\title{ANGLEr: A Next-Generation Natural Language Exploratory Framework}
%
%
\author{Timotej Knez\orcidID{0000-0001-7506-5739} \and
Marko Bajec\orcidID{0000-0002-8502-6765} \and
Slavko Žitnik\orcidID{0000-0003-3452-1106}}
\authorrunning{T. Knez \and M. Bajec \and S. Žitnik}
%
\institute{University of Ljubljana, Ljubljana, Slovenia}
\maketitle              
\begin{abstract}
Natural language processing is used for solving a wide variety of problems. Some scholars and interest groups working with language resources are not well versed in programming, so there is a need for a good graphical framework that allows users to quickly design and test natural language processing pipelines without the need for programming. The existing frameworks do not satisfy all the requirements for such a tool. We, therefore, propose a new framework that provides a simple way for its users to build language processing pipelines. It also allows a simple programming language agnostic way for adding new modules, which will help the adoption by natural language processing developers and researchers. The main parts of the proposed framework consist of (a) a pluggable Docker-based architecture, (b) a general data model, and (c) APIs description along with the graphical user interface. The proposed design is being used for implementation of a new natural language processing framework, called ANGLEr.

\keywords{Natural language processing \and Graphical framework \and Language understanding.}
\end{abstract}
\section{Introduction}
Recently we have seen a lot of interest in the natural language processing tools with Uima~\cite{ferrucci2004uima} receiving around 6 thousand downloads and GATE~\cite{cunningham2002gate} receiving over 400 thousand downloads since 2005\footnote{The number of downloads was recorded by \url{https://sourceforge.net}}. NLP tools are used by people from various backgrounds. For the users outside the computer science area to use the available tools effectively, we have to provide them with an intuitive graphical user interface that allows a simplified construction of text processing pipelines. In this paper, we design an architecture for a new framework that enables quick and simple construction of natural language processing pipelines - ANGLEr. In addition to that, the framework features a way to add new tools that is simple for developers to implement. The researchers could showcase their projects by including them in our framework with very little additional effort. In the past, several frameworks for simplifying NLP workflows were designed. However, they all fall short either in the expandability or the ease of use. In our work we identify the key components of such framework and improve upon the existing frameworks by fixing the identified flaws.

The main contributions of our proposed framework are the following:
\begin{enumerate}
    \item[a)] An architecture that allows simple inclusion of new and already existing tools to the system.
    \item[b)] A data model that is general enough to store information from many different tools, while enabling compatibility between different tools.
    \item[c)] A graphical user interface that speeds up the process of building a pipeline and makes the framework more approachable for users without technical knowledge.
\end{enumerate}

The rest of this paper is organised as follows: in Section~\ref{ch:existing_frameworks} we present the existing frameworks and compare them to our proposed framework. In Section~\ref{ch:angler} we present our framework and its most important parts. We conclude the paper in Section~\ref{ch:conclusion}.

\section{Existing frameworks review}\label{ch:existing_frameworks}
We examined the existing frameworks and selected the ones that we believe to be most useful for users with limited programming knowledge. Analyzing the strengths and weaknesses of the existing tools helps us to define a list of features that could be improved by introducing a new framework. The comparison is summarized in Table~\ref{tab:comparison}. We compare the frameworks on a set of features that were presented by some of the frameworks as their main selling points while others were selected as we find them important for usability. We compare the tools on (1) their graphical user interface (GUI), which affects usability for new users, (2) data model, which affects expandability, and (3) the language required for creation of plugins, which is important for plugin developers.

\textbf{GATE}: One of the best-known frameworks for text processing is GATE~\cite{cunningham2002gate}. It was designed to feature a unified data model that supports a wide variety of language processing tools. While the GATE framework provides a graphical way for pipeline construction, its interface has not been significantly updated in over 10 years. The program is thus difficult to use for new users. Another limitation of the GATE framework is that it requires all of its plugins to be written in the Java language. Because of that, it is difficult to adapt an already existing text processing algorithm to work with the framework.

\textbf{UIMA}: Uima~\cite{ferrucci2004uima} is a framework for extracting information from unstructured documents like text, images, emails and so on. It features a data model that combines extracted information from all previous components. While Uima provides some graphical tools, they are not combined into a single application and do not provide an easy way for creating processing pipelines. The primary way of using UIMA still requires the user to edit XML descriptor documents, which limits usability for new users and slows down the development. New components for the framework can be written in the Java or the C++ programming languages.

\textbf{Orange}: Another popular program for graphical creation of machine learning pipelines is Orange~\cite{demvsar2013orange}. The main feature of the orange framework is a great user interface that is friendly even for non technical users. Orange provides a variety of widgets designed for machine learning tasks and data visualisation. The framework was designed primarily to work with relational data for classic machine learning. It supports two extensions for processing natural language which allow us to use the existing machine learning tools on text documents. The largest drawback when using Orange for text processing is that the tabular representation, used for representing orange data, is not well suited for representing information needed for text processing. The two extensions tackle this problem in different ways.

\textbf{Orange text mining}: The Orange text mining extension implements a data model based on the tabular representation used by the existing tools. This representation allows integration with the existing machine learning and visualization tools, however, it also limits the implementation of some text processing algorithms. Its biggest limitation is that it only works with features on a document level.

\textbf{Textable}: On the other hand, the Textable extension uses a significantly different approach to data representation. In Textable,  different tools produce different data models. This allows for more flexibility when developing new tools, however, it also limits the compatibility between different types of tools. Ideally, we would want a single data model that is capable of supporting every tool. This way we could reach high compatibility between tools while keeping the flexibility when creating new ones.

\textbf{Libraries}: In the past multiple libraries that support text processing have been developed. These libraries are commonly used by experts in the field of natural language processing. They are not well suited for use by other people that might also be interested in language processing. One of such libraries is called OpenNLP~\cite{OpenNLPWebsite}. It supports the development of natural language applications in the Java programming language. There are also multiple Python libraries, such as NLTK~\cite{bird2004nltk}, Gobbli~\cite{nance2021gobbli}, and Stanza~\cite{qi2020stanza}. To use any of these libraries, the user is required to have some programming knowledge. Writing a program is also a lot slower than constructing a pipeline through a graphical interface.

\begin{table}[]
\caption{\label{tab:comparison}Comparison of different natural language processing frameworks.}
\begin{tabular}{l|l|l|l}
\hline
\textbf{Framework} & \textbf{Graphical UI} & \textbf{Unified data model} & \textbf{Plugin language} \\ \hline
GATE                & Yes (Native)  & Yes (too general) & Java \\
UIMA                & Yes (Limited)  & Yes & Java or C++ \\
Orange              & Yes (Native)  & Diverse & Python \\
Orange textable     & Yes (Native)  & Diverse & Python \\
Orange text mining  & Yes (Native)  & Tabular data only & Python \\
Libraries           & No            & Diverse         & Java \\\hline
\bf ANGLEr          & \bf Yes (Web)     & \bf Yes (versioned)  & \bf N/A (Docker packaged) \\\hline
\end{tabular}
\end{table}

\section{ANGLEr: A Next-Generation Natural Language Exploratory Framework}\label{ch:angler}
Based on the problems we identified during the review of the existing frameworks we identified the following important elements: (a) common and versioned data model, (b) extensible Docker-based architecture with API-based communication, and (c) unified and pluggable user interface (see Table~\ref{tab:comparison}).
The data model and APIs will ease the creation of new functionalities for the community. On the other hand, the user interface and simple installation or public hosting are important for the general audience who need to use the language processing tools but do not have the programming skills.

\subsection{Data model}\label{ch:tools}
\begin{figure}[h]
    \begin{center}
        \includegraphics[width=\linewidth]{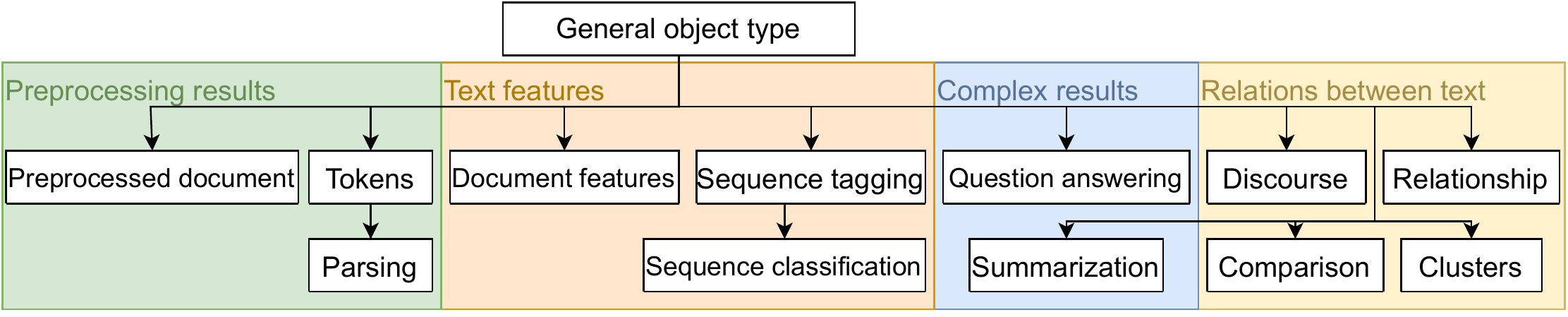}
    \end{center}
\caption{The upper-level hierarchy of ANGLEr object types.}
\label{fig:hierarchy}
\end{figure}

The data model defines a structure that is used to transfer the analysis results between the tools. We aim to support all text processing tools that are available in the existing frameworks as well as new types of algorithms that have not been identified yet. A general view of the data model is presented in Figure~\ref{fig:hierarchy}.

The data model is a central part of the proposed system. It has to provide enough versatility to support every processing tool while keeping a well defined structure to enable compatibility between different tools. We propose a model comprised of two parts. The first part is a corpus of all documents that we want to process. We store the basic metadata about the corpus and each of the documents captured in it. Each document can also be separated into sentences and tokens in case this information was provided in the loaded corpus. The second part of the model stores all of the processor outputs. Outputs and inputs to available processors are encoded in the same schema - i.e., a processor's output can be directly another algorithm's input. We define a list of object types that can be used for representing results as follows:

\begin{description}
\item[Preprocessed document] type stores a new version of the documents in a corpus after a preprocessing algorithm has been applied.
\item[Document features] type stores the values of the features that represent each document in a corpus.
\item[Sequence classification] type is used to store information about the result of the classification of an entire document or a part of a document.
\item[Sequence tagging] type is used to store a set of tags for different parts of the documents. This type could be used for instance to store named entity recognition results.
\item[Relationship] type is used to represent relations between two entities from the source documents.
\item[Discourse] type is used to store entities that appear in the documents. Each entity also contains a list of all entity mentions that correspond to this entity. This object would be used for instance to store coreference information.
\item[Parsing] type stores parts of the document that are linked together. For example, this type would be used to store phrases detected using a chunking algorithm.
\item[Tokens] type stores the information about tokens that appear in the documents. This type would be used to represent for instance the result of a tokenizer or n-gram generator.
\item[Summarization] type is used to store summaries of a document.
\item[Question answering] type is used to store questions and their answers based on the provided text.
\item[Comparison] type is used to store similarity between different parts of the documents.
\item[Clusters] type is used to store clusters of similar documents.
\end{description}

Type hierarchy is proposed to improve backward compatibility when adding new object types. This way each new specific object type contains the attributes of its parent as well as some of its own attributes (each level also allows for key-value metadata storage). An algorithm that was designed to work with a general object type can also work with all of its descendants. For example, a tool for performing named entity recognition would accept the \textit{token} type to get the tokens on which to perform named entity recognition. The user could also provide the \textit{parsing} type since it is a descendant of the \textit{token} type.

\subsection{High-level architecture}

\begin{figure}[htb]
    \begin{center}
        \includegraphics[width=\linewidth]{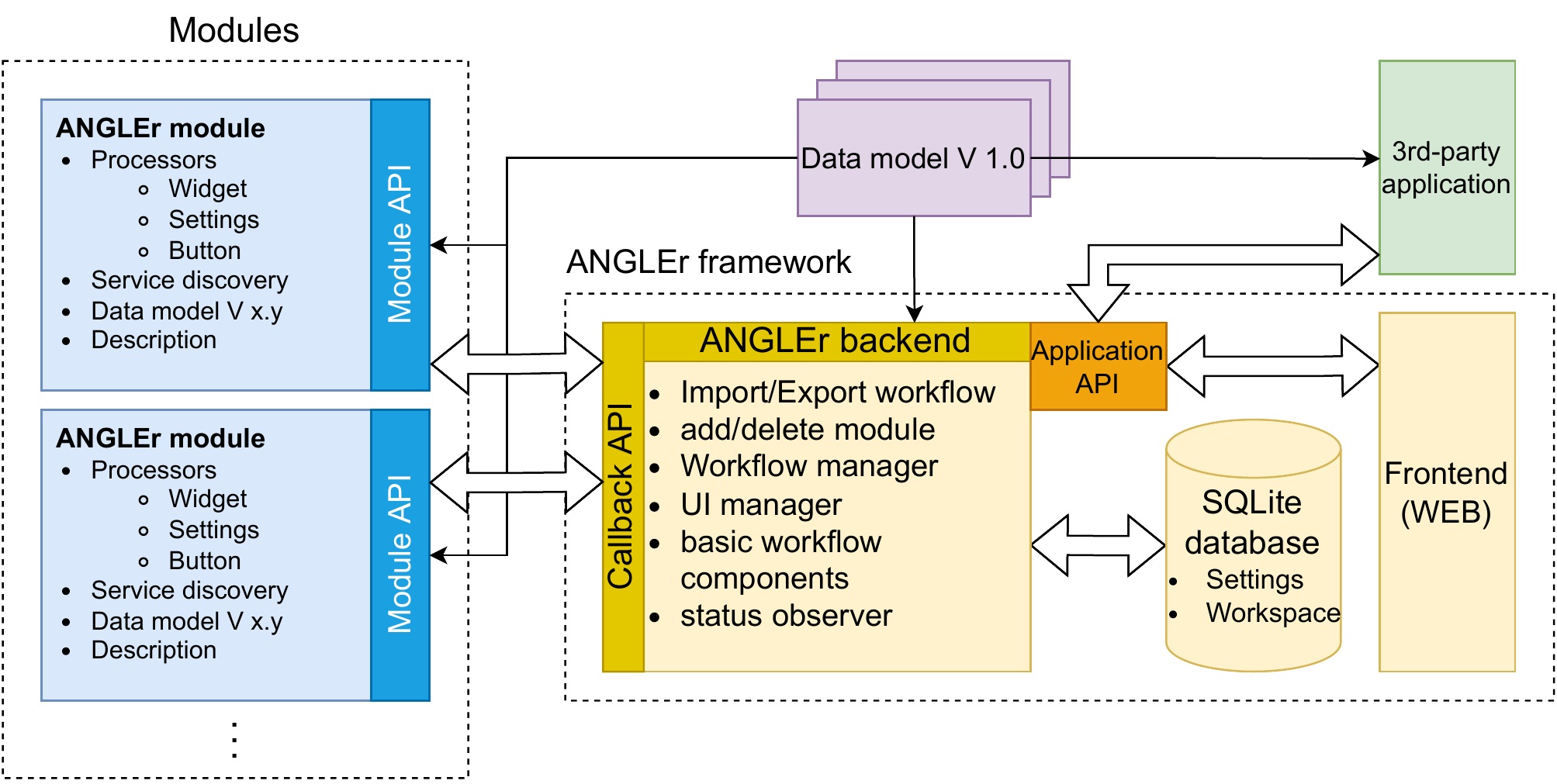}
    \end{center}
\caption{ANGLEr high-level system architecture. The communication between all of the modules an the backend is done using the versioned data model.}
\label{fig:architecture}
\end{figure}

The goal of our framework architecture (presented in Figure~\ref{fig:architecture}) is that adding a new module would be as simple as possible. In addition to that, we want to make sure that the framework and its modules can be also used by third-party applications. Based on this, we have decided that each module should run as a Docker container, which simplifies the inclusion of already existing tools into our framework. The modules are connected to the ANGLEr backend, which is responsible for managing and running the pipeline. The user interacts with the graphical user interface that runs as a web page in a web browser.

\subsection{REST interface}
All of the communication between the parts of the framework is done over REST application programming interfaces (API). This also allows third-party applications to access the modules and the ANGLEr backend. 

\begin{table}[h]
\caption{Endpoints in the \textit{Module API}. The underlined attributes are required. The processors endpoint returns a list of objects where each object has the presented attributes.}
\begin{tabular}{lll}
\textbf{Endpoint} & \textbf{Parameters} & \textbf{Description} \\ \hline
/about & {\ul UUID} & Identifier of the module. \\
 & {\ul name} & Name of the module. \\
 & {\ul version} & Module version. \\
 & {\ul data\_model} & Version of the data model used. \\
 & desc & Module description. \\
 & authors & List of authors. \\
 & organisation & Organisation of the authors. \\
 & url & URL address of the page about the module. \\ \hline
/processors & {\ul name} & Name of a processor. \\
 & short\_name & Short version of the name. \\
 & {\ul data\_endpoint} & Endpoint where the data for processing can be sent. \\
 & settings\_endpoint & An address of a page containing processor settings. \\
 & ui\_endpoint & An address of a page for visualization. \\
 & {\ul icon} & \begin{tabular}[c]{@{}l@{}}An address of the icon to be used to represent the \\ processor.\end{tabular} \\
 & {\ul category} & \begin{tabular}[c]{@{}l@{}}A category of the processors menu that should \\ contain this processor.\end{tabular} \\
 & {\ul inputs} & \begin{tabular}[c]{@{}l@{}}A list of objects representing different inputs. Each \\ object should contain a name and a list of types required.\\ An example is shown in Figure~\ref{fig:input_example}\end{tabular} \\
 & {\ul outputs} & \begin{tabular}[c]{@{}l@{}}A list of objects representing different outputs. Each \\ object should contain a name and a type of the output.\end{tabular} \\ \hline
/docs & html\_page & A web page containing the documentation. \\ \hline
\end{tabular}
\label{tab:apis}
\end{table}

\begin{figure}[h]
    \begin{center}
        \includegraphics[width=\linewidth]{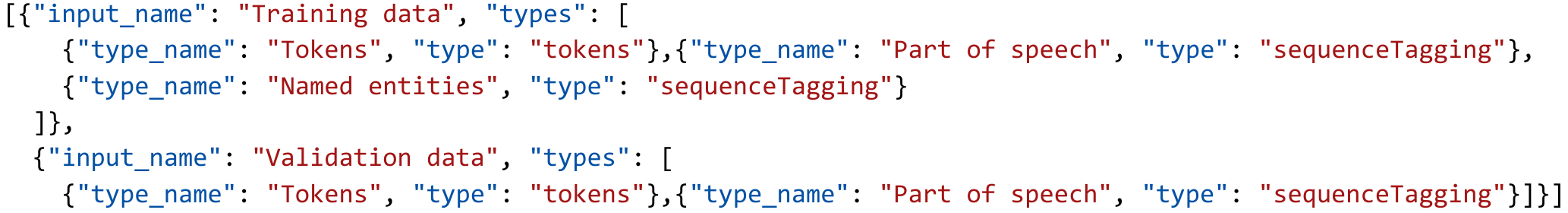}
    \end{center}
\caption{An example of the input attribute for a named entity recognition module. }
\label{fig:input_example}
\end{figure}

The framework features multiple REST APIs shown in Figure~\ref{fig:architecture}. The \textit{Module API} (Table~\ref{tab:apis}) is responsible for commands that are sent from the ANGLEr backend to a module. It allows the ANGLEr backend to gather information about the processors that are running in a module. It also allows the backend to send the data and start its processing. After the processing is done, the module sends the results to the backend using the \textit{Callback API}. The ANGLEr backend also exposes the \textit{Application API}, which is responsible for communication with the graphical user interface, as well as with any third-party applications that might want to use the ANGLEr functionality. Since the \textit{Module API} has to be implemented by each module, we present its endpoints in Table~\ref{tab:apis}.

Once a module is added to the framework, the ANGLEr backend creates a request to the \textit{about} endpoint, which provides the basic information about the entire module. After that, it requests the \textit{processors} endpoint, which lists all processors that are contained in the module. This information allows the framework to visualise the processor in the graphical interface. It also provides the endpoints that the ANGLEr backend uses to send data for processing and to show configuration and visualisation pages.

\subsection{Graphical user interface}

\begin{figure}[h]
    \begin{center}
        \includegraphics[width=1\linewidth]{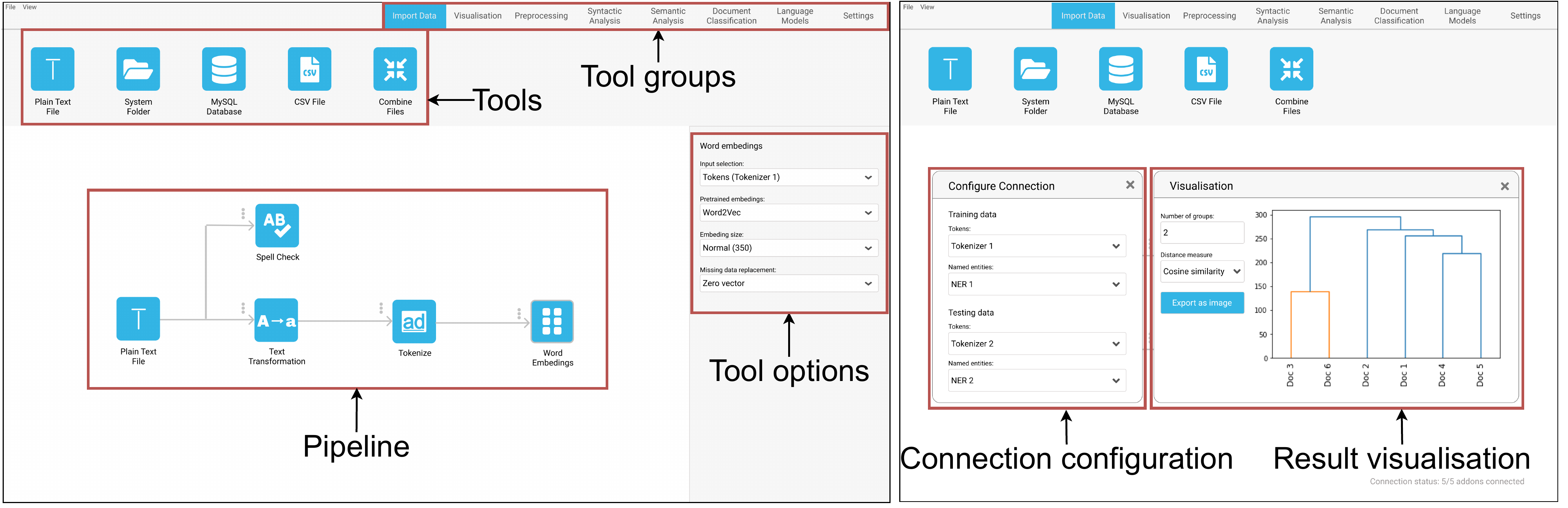}
    \end{center}
\caption{The main parts of the ANGLEr user interface.}
\label{fig:ui}
\end{figure}

The user interface is a key component for making the framework simple to use. It allows the users without any programming knowledge to use the framework. This is especially important since language processing tools are very useful for linguists, who typically do not have advanced computer knowledge. We propose a simple design for the user interface, which is shown in Figure~\ref{fig:ui}. The tabs at the top of the page organise the tools into groups based on their purpose. For example, the groups contain the tools for text preprocessing, for semantic and syntactic analysis etc. A user can get a widget for each tool by dragging and dropping. The widget has a page for setting its parameters and a dialog for selecting the input. A widget can optionally also provide a page for data visualisation (see the Figure~\ref{fig:ui} right). The user connects the widgets into a pipeline that can be stored to a file and loaded by any user. The ANGLEr framework then executes the pipeline to process the data.

\section{Conclusion}\label{ch:conclusion}
We describe the main components for implementing a new natural language processing framework - ANGLEr. We believe that a new framework based on our proposal would provide a large improvement over the existing frameworks and would greatly benefit users that are working with natural language processing. The framework would provide a fast way for prototyping when developing text processing pipelines. It would also allow users with no programming knowledge to build advanced NLP pipelines. In addition to that, the new framework would provide the researchers with a great way for showcasing their work in the NLP area. Since the tools can be implemented in any programming language, their inclusion is much less complicated than with existing frameworks.

\bibliographystyle{splncs04}
\bibliography{bibliography}

\begin{thebibliography}{1}
\providecommand{\url}[1]{\texttt{#1}}
\providecommand{\urlprefix}{URL }
\providecommand{\doi}[1]{https://doi.org/#1}

\bibitem{OpenNLPWebsite}
Apache: Opennlp (2010), \url{http://opennlp.apache.org}

\bibitem{bird2004nltk}
Bird, S., Loper, E.: Nltk: the natural language toolkit. Association for
  Computational Linguistics (2004)

\bibitem{cunningham2002gate}
Cunningham, H.: Gate, a general architecture for text engineering. Computers
  and the Humanities  \textbf{36}(2),  223--254 (2002)

\bibitem{demvsar2013orange}
Dem{\v{s}}ar, J., Curk, T., Erjavec, A., Gorup, {\v{C}}., Ho{\v{c}}evar, T.,
  Milutinovi{\v{c}}, M., Mo{\v{z}}ina, M., Polajnar, M., Toplak, M.,
  Stari{\v{c}}, A., et~al.: Orange: data mining toolbox in python. the Journal
  of machine Learning research  \textbf{14}(1),  2349--2353 (2013)

\bibitem{ferrucci2004uima}
Ferrucci, D., Lally, A.: Uima: an architectural approach to unstructured
  information processing in the corporate research environment. Natural
  Language Engineering  \textbf{10}(3-4),  327--348 (2004)

\bibitem{nance2021gobbli}
Nance, J., Baumgartner, P.: gobbli: A uniform interface to deep learning for
  text in python. Journal of Open Source Software  \textbf{6}(62), ~2395 (2021)

\bibitem{qi2020stanza}
Qi, P., Zhang, Y., Zhang, Y., Bolton, J., Manning, C.D.: Stanza: A python
  natural language processing toolkit for many human languages. In: Proceedings
  of the 58th Annual Meeting of the Association for Computational Linguistics:
  System Demonstrations. pp. 101--108 (2020)

\end{thebibliography}
\end{document}